\documentclass[10pt,twocolumn,letterpaper]{article}

\usepackage[pagenumbers]{cvpr} 

\usepackage{graphicx}
\usepackage{amsmath}
\usepackage{amssymb}
\usepackage{booktabs}

\usepackage{multirow}        
\usepackage{adjustbox}       
\usepackage{caption}
\usepackage{subcaption}
\usepackage{amsmath}

\usepackage[pagebackref,breaklinks,colorlinks]{hyperref}

\usepackage[capitalize]{cleveref}
\crefname{section}{Sec.}{Secs.}
\Crefname{section}{Section}{Sections}
\Crefname{table}{Table}{Tables}
\crefname{table}{Tab.}{Tabs.}

\def\cvprPaperID{10884} 
\def\confName{CVPR}
\def\confYear{2022}

\begin{document}

\title{Norm-Scaling for Out-of-Distribution Detection}

\author{Deepak Ravikumar,  Kaushik Roy\\
School of Electrical and Computer Engineering,\\ 
Purdue University, West Lafayette, IN 47907, USA\\

{\tt\small \{dravikum, kaushik\}@purdue.edu}
}
\maketitle

\begin{abstract}
Out-of-Distribution (OoD) inputs are examples that do not belong to the true underlying distribution of the dataset. Research has shown that deep neural nets make confident mispredictions on OoD inputs. Therefore, it is critical to identify these OoD inputs for safe and reliable deployment of deep neural nets. Often a threshold is applied on a similarity score to detect OoD inputs. One such similarity is angular similarity which is the dot product of latent representation with the mean class representation. Angular similarity encodes uncertainty, for example, if the angular similarity is less, it is less certain that the input belongs to that class.  However, we observe that, different classes have different distributions of angular similarity. Therefore, applying a single threshold for all classes is not ideal since the same similarity score represents different uncertainties for different classes. In this paper, we propose norm-scaling which normalizes the logits separately for each class. This ensures that a single value consistently represents similar uncertainty for various classes. We show that norm-scaling, when used with maximum softmax probability detector, achieves \textbf{9.78\%} improvement in AUROC, \textbf{5.99\%} improvement in AUPR and \textbf{33.19\%} reduction in FPR95 metrics over previous state-of-the-art methods.
\end{abstract}


\section{Introduction}
Deep learning models deployed in the real world often encounter inputs that are unlike the training set. However, they may erroneously classify these inputs with very high confidence \cite{goodfellow2014explaining, nguyen2015deep}.
It is critical to identify and flag such Out-of-Distribution (OoD) inputs to enable reliable and safe deployment of applications such as bacterial identification based on genomic sequence \cite{ren2019likelihood}, self driving cars \cite{papernot2017practical}, medical diagnosis \cite{shen2017deep} and other safety critical applications \cite{amodei2016concrete}.

Various approaches have been proposed in literature to identify OoD examples, such as Mahalanobis distance based detection \cite{NIPS2018mahalanobis}, Generative Adversarial Networks (GANs) based methods \cite{deecke2018image, lee2018training, ren2019likelihood}, energy score based approaches \cite{liu2020energybased} and softmax confidence based techniques \cite{hendrycks17baseline, hendrycks2019oe, Hsu_2020_CVPR, liang2020enhancing}. 
Most approaches can be interpreted as using the angular similarity (or a proxy for angular similarity) to detect OoD examples. Angular similarity is the dot product between the latent representation and the mean representation for a specific class.  A threshold is applied on the angular similarity (or a proxy) to identify OoD examples. Proxies such as softmax score \cite{hendrycks17baseline, hendrycks2019oe} and energy score \cite{liu2020energybased} have been shown to be very successful at detecting OoD examples.

Angular similarity encodes uncertainty. For example, large angular similarity implies more certainty that the input belongs to that class, while small angular similarity suggests the opposite. We observe that the angular similarity has different distributions for different classes. This can be naively observed as different distribution means. The logit (or activation) value of the final classification layer corresponds to the angular similarity of the latent representation to the mean class representation. This is because the weights of the classification layer represent  the average class representation. Hence, the dot product of weight with the layer input represents the angular similarity. 

\begin{figure}[hbt!]
    \centering
    \includegraphics[scale=0.36]{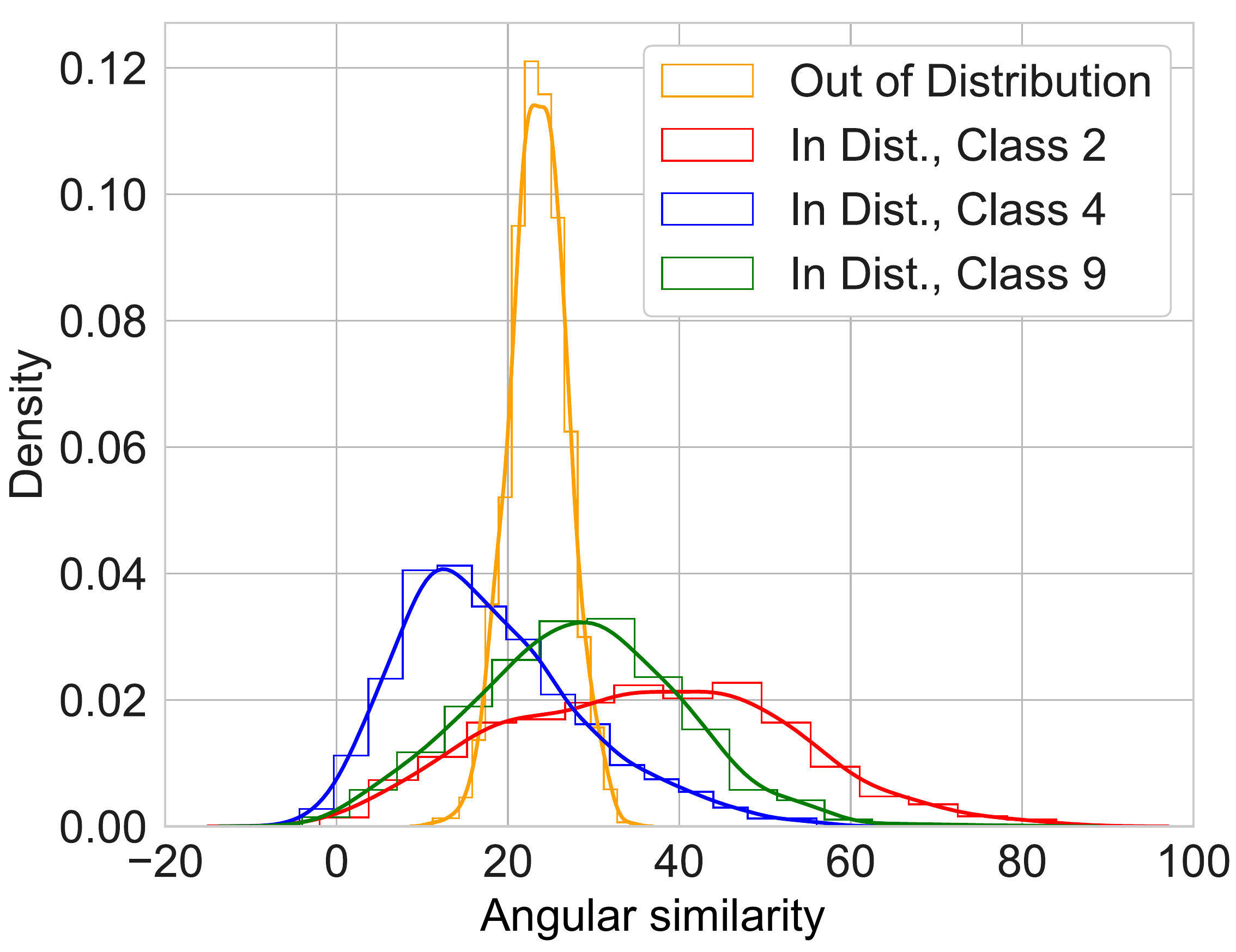}
    \caption{Probability density (Kernel Density Estimate, solid) and probability mass (stepped) for angular similarity
    of classes 2, 4 and 9 of CIFAR-10 dataset with Gaussian Noise as OoD dataset for a ResNet18 network.}
    \label{fig:diff_dis}
\end{figure}

We  show the probability mass function of the angular similarity  obtained from ResNet18 \cite{he2016deep} trained on CIFAR-10 \cite{krizhevsky2009learning} in \cref{fig:diff_dis}. It visualizes the kernel density estimate (thick line) and the empirical probability mass function (stepped line) of the angular similarity for classes 2, 4 and 9, respectively and the angular similarity of OoD (Gaussian Noise) inputs with the predicted class. To obtain the mass function for classes 2, 4 and 9 we infer on the in-distribution data, and plot the mass for the corresponding logit when the predicted class is 2, 4 and 9 respectively. Similarly for OoD data we infer on the OoD samples and plot the distribution for the logits of the predicted class. The logit values as shown in \cref{fig:diff_dis} are threshold-ed in case of MSP detector \cite{hendrycks17baseline} to detect OoD examples. 

Observing the plot from \cref{fig:diff_dis} we see that different classes have different distributions. The uncertainty can be approximated using the density estimate. For example if the input is predicted to be class 4 and the angular similarity is 20, it is very certain (density $\sim$ 0.038) that this input is truly class 4. On the other hand, if the input is predicted as class 2 the same angular similarity of 20 represents much lower certainty (density $\sim$ 0.018) of the input being class 2. Therefore, an angular similarity  $s$ (or its corresponding proxy $p$) represents different uncertainties for different classes.  This suggests that applying a single threshold on $s$ (or $p$) for various classes reduces detection performance. For example, a threshold at 20 may separate class 4 from OoD but this threshold does not separate classes 2 and 9 from OoD. 

In this paper, we propose norm-scaling that normalizes the logits using Z-score normalization. The normalization is applied separately for each class. In other words, each class has a corresponding mean and standard deviation which is used to normalize the logit corresponding to that class. Such a normalization scheme ensures that  similarity $s$ (or a proxy $p$) represents similar uncertainty across different classes.  The proposed norm-scaling of the logits when applied along maximum softmax probability detector \cite{hendrycks17baseline} achieves significant improvement in AUROC, AUPR and FPR95 metrics over previous state-of-the-art methods. Further, we show that norm-scaling performs better than applying a separate threshold for each class. We also provide an alternate perspective on norm-scaling. We show that norm-scaling can be viewed as a parameter free version of temperature scaling. Temperature scaling is a technique \cite{platt1999probabilistic} that divides the logits by a temperature (hyper)parameter and has been shown to improve OoD detection \cite{guo2017calibration, liang2020enhancing}.

In summary the contributions of this paper are:
\begin{itemize}
    \item We observe that a single value of angular similarity $s$ (or a proxy $p$) represents different uncertainties for different classes.
    \item We propose norm-scaling, a normalization scheme which ensures consistent uncertainty values for different classes. We show that it is better than the approach of using a threshold for each class. Further, we show that norm-scaling can be viewed as a parameter free version of temperature scaling.
    \item We show that norm-scaling applied on maximum softmax probability detector \cite{hendrycks17baseline} achieves 9.78\% improvement in AUROC, 5.99\% improvement in AUPR and 33.19\% reduction in FPR95 metrics over previous state-of-the-art methods.
\end{itemize}

\section{Related Work}
There have been many approaches in literature that have attempted to address the challenge of OoD detection. Research \cite{guo2017calibration, kuleshov2018accurate, maddox2019simple} has shown that well calibrated scores can improve OoD detection performance. Temperature scaling \cite{guo2017calibration} has been shown to improve calibration. It was leveraged by the authors of ODIN \cite{Hsu_2020_CVPR, liang2020enhancing} to improve upon the OoD detection performance of Maximum Softmax Probability (MSP) detector. MSP detector \cite{hendrycks17baseline} on the other hand used un-calibrated softmax scores for OoD detection.
Another interesting approach is the use of mixup \cite{zhang2018mixup, thulasidasan2019mixup}
to improve OoD detection performance. Authors of \cite{thulasidasan2019mixup} find that mixup trained networks are significantly better calibrated and are less prone to over-confident predictions on out-of-distribution and random-noise data.

Recent research  \cite{liu2020energybased} has also argued for the use of energy score for OoD detection which is theoretically more aligned with the probability density of the inputs and is therefore less likely to result in overconfident predictions. Other approaches include modeling the underlying in-distribution dataset using generative models such as Gaussian Discriminant Analysis \cite{NIPS2018mahalanobis} or Generative Adversarial Networks (GANs) \cite{gans, lee2018training, ren2019likelihood} to separate in-distribution examples from out-of-distribution examples.

Further, it has also been shown that incorporating an auxiliary OoD set during training \cite{hendrycks2019oe, liu2020energybased}  improves  OoD detection performance. These techniques are often categorized under supervised OoD detection as opposed to unsupervised techniques that do not use auxiliary OoD datasets. This means that supervised techniques assume a prior on the OoD dataset in the form of the auxiliary OoD set. Some works \cite{liang2020enhancing} have suggested that it is very hard to define such priors. In \cite{hendrycks2019oe} the authors extended a previous work \cite{hendrycks17baseline} by modifying the loss function used to train the classifier. They trained the network on an in-distribution dataset as well as an auxiliary outlier (OoD) dataset. The authors claim that the proposed method of exposing the network to outliers enables the detectors to generalize better and detect unseen anomalies.
Most of these previously described approaches  can be viewed as applying a threshold on the angular similarity or a proxy (such as softmax confidence, energy score etc.) to detect OoD examples. We observe that a single threshold when applied for different classes reduces OoD detection performance as the same similarity (or proxy) value represents different uncertainties for different classes. 
To address this issue, we propose norm-scaling.

\section{Methodology}
The proposed OoD detection technique is detailed in the following subsections.
\subsection{Norm-Scaling}

Let us consider the proposed norm-scaling, a novel scaling technique that performs Z-score normalization on the logits prior to softmax. Z-score normalization is used to address the issue of inconsistent uncertainty representation between classes. 
Mathematically norm-scaling can be described by the following set of equations
\begin{equation}
    \mu_j^{tr} = \frac{1}{D^{tr}}\sum_{k=1}^{D^{tr}}z_{kj}
\label{eqn:mean_logit}
\end{equation}
\begin{equation}
    \sigma_j^{tr} = \sqrt{\frac{1}{D^{tr}}\sum_{k=1}^{D^{tr}}(z_{kj} - \mu_j)^2}
\label{eqn:std_logit}
\end{equation}
\begin{equation}
    \begin{array}{cl}
    z^{s}_{ij} = \cfrac{z_{ij} - \mu_j}{\sigma_j} & j \in {1, 2, .., N}
    \end{array}
\label{eqn:norm_scaling}
\end{equation}

where $z^{s}_{ij}$ and $z_{ij}$ are the norm-scaled and un-scaled logits respectively, for the $j^{th}$ output class and $i^{th}$ image in the training dataset of size $D^{tr}$ and $N$ is the number of classes. 
The norm-scaled logits $z^{s}_{ij}$ are used to compute the softmax scores, which forms the basis of the maximum softmax probability OoD detector as described in \cite{hendrycks17baseline}. The maximum softmax probability detector applies a threshold on the softmax confidence of the predicted class to detect OoD examples. That is, inputs whose softmax confidence are less than a threshold value are considered as OoD inputs.

During testing we may or may not have batches for the dataset to calculate the mean and standard deviation. Therefore we assume we get one example at a time $t$ and use a running mean and average for the logits of each class as described by the following equations
\begin{equation}
    \begin{array}{cl}
    \mu_j^{t} = \cfrac{\mu_j^{t-1} +  z_{j}}{t + 1}
     &  t \in {1, 2, .., T}
        \end{array}
\end{equation}
\begin{equation}
    \mu_j^{0} = \mu_j^{tr}
\end{equation}
\begin{equation}
    (\sigma_j^{t})^{2} = \frac{(\sigma_j^{t-1})^{2} +  (z_{j} - \mu_j^{t})^{2}}{t + 1} \\
\end{equation}
\begin{equation}
    (\sigma_j^{0})^{2} = (\sigma_j^{tr})^{2}
\end{equation}
where $\mu_j^{tr}$, $\sigma_j^{tr}$ are the mean and standard obtained from the training set described in  \cref{eqn:mean_logit} and \cref{eqn:std_logit}, $\mu_j^{t}$ and $\sigma_j^{t}$  are the running mean and standard deviation for the $j^{th}$ output class at time $t$. Thus, during testing we use $\mu_j^{t}$ and $\sigma_j^{t}$ in \cref{eqn:norm_scaling} to perform norm-scaling.


\begin{figure}[htp!]
        \centering
        \includegraphics[scale=0.37]{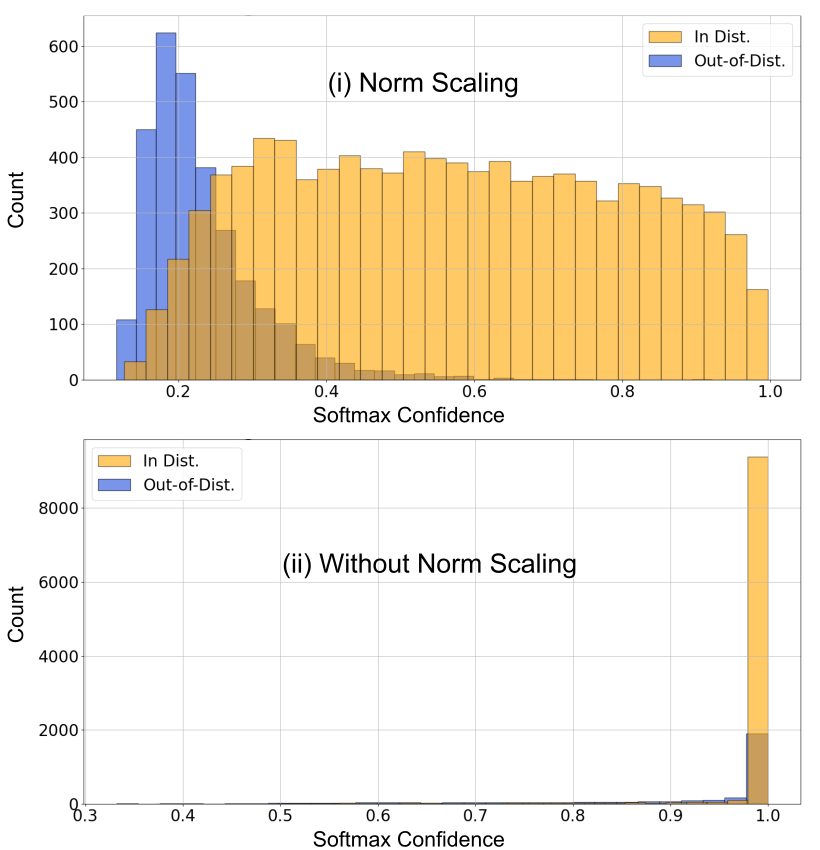}
        \caption{Softmax histograms for a  ResNet-18 trained on CIFAR-10. Effect of using (i) norm-scaling, (ii) without norm-scaling. The histograms use the same model, trained on CIFAR-10 with LSUN as OoD dataset.}
        \label{fig:effect_of_norm_scaling}
\end{figure}

\cref{fig:effect_of_norm_scaling} shows the effect of norm-scaling on the softmax confidence score of predicted class for in-distribution and out-of-distribution datasets. From  \cref{fig:effect_of_norm_scaling} we can clearly see that in the absence of norm-scaling (refer to (ii) in \cref{fig:effect_of_norm_scaling}) the two distributions overlap significantly reducing separability. The use of norm-scaling (refer to (i) in \cref{fig:effect_of_norm_scaling}) improves the separability of of the two distributions making OoD detection easier and more effective. We perform more extensive experiments to validate this observation as detailed in the \emph{Experiments} section.

\section{Experiments}
\label{sec:exp}
This section consists of three parts:  the first part introduces the terminology and metrics used for performance measurement. Second, compares the OoD detection performance of norm-scaling with other methods, and finally we provide a temperature scaling perspective of norm-scaling.

\subsection{Terminology and Metrics}
This section provides a brief description of the terminology and metrics used in the paper.

\textbf{Calibration} The goal of calibration is to ensure that model confidence reflects the ground truth correctness likelihood. Common calibration techniques include, histogram binning \cite{zadrozny2001obtaining}, isotonic regression \cite{zadrozny2002transforming}, Bayesian binning \cite{naeini2015obtaining} and Platt Scaling \cite{platt1999probabilistic, niculescu2005predicting}.

\begin{figure}[hbt!]
    \centering
    \includegraphics[scale=0.51]{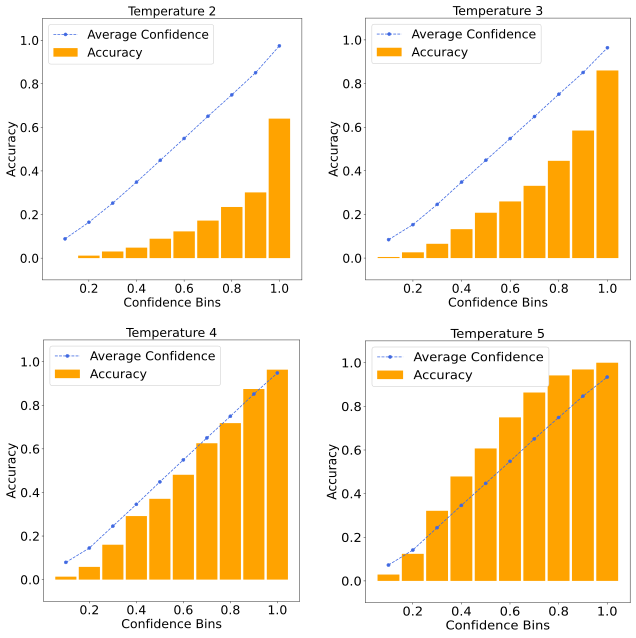}
    \caption{Reliability diagrams for a CIFAR-100 trained ResNet-18 at various scaling temperatures.}
    \label{fig:rd_ex}
\end{figure}

\textbf{Reliability Diagrams} are tools to visualize model calibration \cite{degroot1983comparison, niculescu2005predicting}. These diagrams plot empirical sample accuracy as a function of confidence.  \cref{fig:rd_ex} shows  reliability diagrams for the same ResNet-18 model trained on CIFAR-100 \cite{krizhevsky2009learning} at various temperatures.  \cref{fig:rd_ex} was generated by splitting the confidence range $[0, 1]$ into $M$ bins. All inputs whose predicted confidence falls in the interval $(\frac{m-1}{M}, \frac{m}{M}]$ is assigned to bin $B_{m}$ where $B_{m}$ is the $m^{th}$ bin. The accuracy (orange bars in  \cref{fig:rd_ex}) corresponding to bin $B_{m}$ is given by
\begin{equation}
    acc(B_{m}) = \frac{1}{|B_{m}|}\sum_{i\in B_{m}}^{}[\hat{y_{i}} = y_{i}]
\label{eqn:bin_acc}
\end{equation}
where $[$ $]$ is the Iverson bracket notation for the Kronecker delta function, $\hat{y_{i}}$ is the ground truth label and $y_{i}$ is the predicted label for the $i^{th}$ sample in bin $m$. The average confidence for the bin $B_{m}$ (the blue line in \cref{fig:rd_ex}) is given by
\begin{equation}
    conf(B_{m}) = \frac{1}{|B_{m}|}\sum_{i\in B_{m}}^{}p_{i}
\label{eqn:bin_conf}
\end{equation}
where $p_{i}$ is the confidence predicted by the model for the $i^{th}$ sample in bin $m$. For an ideally calibrated model,  the empirical accuracy for each bin would be identical to the average predicted model confidence i.e. $conf(B_{m}) = acc(B_{m})$ $\forall$ $m = 1, 2, ..., M$. This ideal temperature is close to 4 for ResNet-18 trained on CIFAR-100 as seen in \cref{fig:rd_ex} where the blue plot shows the ideal accuracy (equal to average confidence) for the corresponding confidence bins.

\textbf{Expected Calibration Error (ECE)} Reliability diagrams do not account for the proportion of the samples in each bin, therefore they cannot be used to calibrate the model. This is addressed by Expected Calibration Error \cite{naeini2015obtaining} which is given by
\begin{equation}
    ECE = \sum_{m=1}^{M}\frac{|B_{m}|}{n}\left|acc(B_{m}) - conf(B_{m}) \right|
\label{eqn:ece}
\end{equation}
The lower the expected calibration error better the calibration of the model.



\textbf{Recall, FPR, Precision} The performance of a binary classification algorithm is often evaluated using Recall or True Positive Rate (TPR), False Positive Rate (FPR) and Precision. Recall (TPR), FPR and Precision are defined by  \cref{eqn:TPR}, \cref{eqn:FPR} and \cref{eqn:precision} respectively.

\begin{equation}\label{eqn:TPR}
    \begin{array}{cc}
      Recall = TPR = \cfrac{TP}{TP+FN}
    \end{array}
\end{equation}
\begin{equation}\label{eqn:FPR}
    \begin{array}{cc}
         FPR = \cfrac{FP}{FP+TN}
    \end{array}
\end{equation}
\begin{equation}\label{eqn:precision}
    \begin{array}{cc}
         Precision = \cfrac{TP}{TP+FP}
    \end{array}
\end{equation}
where, TP is True Positive, FP is False Positive, TN is True Negative and FN is False Negative.

\textbf{AUROC} The Receiver Operating Characteristic (ROC) is a plot of the True Positive Rate (TPR) against the False Positive Rate (FPR). The area under this ROC plot is referred to as the Area Under the Receiver Operating Characteristic AUROC. An AUROC of 1 denotes an ideal detection scheme. 

\textbf{AUPR} The area under the Precision Recall plot is the AUPR. Similar to AUROC an AUPR of 1 denotes an ideal detection scheme. 
    
\textbf{FPR at TPR of 95\% (FPR95)} Denotes the False Positive Rate (FPR) when the True Positive Rate (TPR) is 95\%. Lower value of FPR at TPR of 95\% indicates a better classifier.


{\renewcommand{\arraystretch}{1.7}
\begin{table*}[!hbt]
\begin{adjustbox}{width=\textwidth}
\begin{tabular}{c|ccc|ccc|ccc}
\hline
\multirow{4}{*}{In-Dataset}             
& \multicolumn{3}{c|}{\multirow{2}{*}{AUROC $\uparrow$}} 
& \multicolumn{3}{c|}{\multirow{2}{*}{AUPR $\uparrow$}}  
& \multicolumn{3}{c}{\multirow{2}{*}{FPR95 $\downarrow$}} \\

& \multicolumn{3}{c|}{}
& \multicolumn{3}{c|}{}
& \multicolumn{3}{c}{} \\  \cline{2-10} 
& \multirow{2}{*}{Baseline \cite{hendrycks17baseline}} & \multirow{2}{*}{Multi-Thresh.} & \multirow{2}{*}{Norm-Scaling} 
& \multirow{2}{*}{Baseline \cite{hendrycks17baseline}} & \multirow{2}{*}{Multi-Thresh.} & \multirow{2}{*}{Norm-Scaling} 
& \multirow{2}{*}{Baseline \cite{hendrycks17baseline}} & \multirow{2}{*}{Multi-Thresh.} & \multirow{2}{*}{Norm-Scaling} \\
             &                       &                      &                                    &                       &                      &                            &                      &                &             \\ \hline
CIFAR-10     & 0.8234 $\pm$ 0.0538 	 & 0.8681 $\pm$ 0.0379  & \textbf { 0.9392 $\pm$ 0.0081 } 	 & 0.7803 $\pm$ 0.1113 	 & 0.8661 $\pm$ 0.0314   & \textbf { 0.9109 $\pm$ 0.0115 } 	 & 0.6079 $\pm$ 0.1106 	& 0.5414 $\pm$ 0.0657  & \textbf { 0.2160 $\pm$ 0.0236 } \\ 
CIFAR-100    & 0.6645 $\pm$ 0.1680 	 & 0.7542 $\pm$ 0.0187  & \textbf { 0.8931 $\pm$ 0.0027 } 	 & 0.6204 $\pm$ 0.1990 	 & 0.7580 $\pm$ 0.0241  & \textbf { 0.8397 $\pm$ 0.0036 } 	 & 0.6577 $\pm$ 0.1304 	 &  0.7683 $\pm$ 0.0290 & \textbf { 0.3305 $\pm$ 0.0069 } \\ 
SVHN         &  0.9235 $\pm$ 0.0074  & \textbf {0.9398 $\pm$ 0.0031} & 0.9181 $\pm$ 0.0064 	     & 0.7610 $\pm$ 0.1088   & \textbf{0.9740 $\pm$ 0.0019}   & 0.7760 $\pm$ 0.0183 	 & \textbf { 0.2660 $\pm$ 0.0442 } 	& 0.3745 $\pm$ 0.0148  & 0.3395 $\pm$ 0.0291 \\ 
TinyImageNet & 0.5510 $\pm$ 0.1104 	 & 0.6613 $\pm$ 0.0212  & \textbf { 0.7938 $\pm$ 0.0066 } 	 & 0.5272 $\pm$ 0.1990 	 & 0.7280 $\pm$ 0.0159   & \textbf { 0.7457 $\pm$ 0.0068 } 	 & 0.8261 $\pm$ 0.0658 	 & 0.8972 $\pm$ 0.0244  & \textbf { 0.5010 $\pm$ 0.0140 } \\ \hline
\end{tabular}
\end{adjustbox}
\caption{Comparing performance of Maximum Softmax Probability (MSP) detector, MSP with multiple thresholds and MSP with  norm-scaling (averaged over 5 different seeds, mean $\pm$ std) on various OoD datasets (Uniform Noise, Gaussian, SVHN, Textures, LSUN, TinyImageNet, CIFAR-10 and CIFAR-100, Places365). Expanded version of the table is available in the Appendix.}
\label{tab:norm_scaling_ood_perf}
\end{table*}}

\subsection{OoD Performance}
This subsection details the experimental setup used for evaluating the OoD performance of the proposed detector. It also details the results of the experiment and compares the performance of the proposed method with various state-of-the-art OoD detection schemes.
\subsubsection{Setup}
\label{sec:exp_setup}
To evaluate the OoD performance, we train a network on an (in-distribution) dataset, and use the proposed detector during testing on various OoD datasets. This setup is standard practice in literature \cite{hendrycks17baseline, liang2020enhancing, liu2020energybased, ren2019likelihood, NIPS2018mahalanobis}. 

We use the datasets listed in \cref{tab:dataset_sizes} as OoD datasets. Note, when using a dataset as in-distribution we exclude it from the OoD list. For example, when we train a network on SVHN (in-distribution) we use Gaussian Noise, Uniform Noise, CIFAR-10, CIFAR-100, TinyImageNet, LSUN, Textures and Places365 as OoD datasets to evaluate OoD detection performance. 

We report the results for in-distribution datasets (eg. SVHN) by averaging AUROC, AUPR and FPR95 metrics obtained on each OoD dataset. Further to improve confidence in the results we train 5 differently seeded models and obtain mean and standard deviations for the metrics. We test using  CIFAR-10, CIFAR-100, SVHN and TinyImageNet as in-distribution datasets. All the tables i.e. \cref{tab:norm_scaling_ood_perf} and \cref{tab:comp_other} follow this reporting scheme described above.

{\renewcommand{\arraystretch}{1.2}
\begin{table}[hbt!]
\centering
\begin{adjustbox}{}
\begin{tabular}{c|c}
\hline
Dataset Name   & Testset Size             \\ \hline
CIFAR-100 \cite{krizhevsky2009learning}      & 10000                    \\ 
TinyImageNet \cite{tinyimagenet}   & 10000                    \\
LSUN  \cite{yu2015lsun}         & 3000                     \\
Places365 \cite{zhou2017places}    & 36500                    \\
Textures \cite{cimpoi14describing}      & 5640                     \\
SVHN  \cite{svhn}         & 26032                    \\
Gaussian Noise & Size of In-Dist. testset \\ 
Uniform Noise  & Size of In-Dist. testset \\ \hline
\end{tabular}
\end{adjustbox}
\vspace{2mm}

\caption{OoD Datasets used and their sizes. }
\label{tab:dataset_sizes}

\end{table}}

{\renewcommand{\arraystretch}{1.3}
\begin{table}[hbt!]
\centering
\begin{adjustbox}{}
\begin{tabular}{c|c}
\hline
Dataset                       & Accuracy           \\ \cline{1-2}
CIFAR-10                      & 92.90 $\pm$ 0.27   \\
CIFAR-100                     & 71.62 $\pm$ 0.18   \\
SVHN                          & 95.69 $\pm$ 0.16   \\
TinyImageNet                  & 32.80 $\pm$ 0.17   \\ \hline
\end{tabular}
\end{adjustbox}
\vspace{2mm}

\caption{Baseline accuracies (mean $\pm$ std. averaged over 5 seeds) of ResNet-18 trained models on various datasets.}
\label{tab:baseline_cln_acc}

\end{table}}

We use a ResNet18 \cite{he2016deep} model trained until convergence and the baseline accuracies of the networks are show in  \cref{tab:baseline_cln_acc}. The training procedure used the SGD optimizer with a momentum of 0.9 and weight decay of $5\times10^{-4}$. The training used a 90\%-10\% training-validation split with the initial learning rate set to $10^{-2}$ and it was scaled down by a factor of 10 at 60\% and 80\% completion using a learning rate scheduler. All the experiments were carried out on a single Nvidia Titan Xp GPU and Intel 10700KF CPU with 16 GB of RAM. All of the metrics (accuracy, AUROC, AUPR and FPR95) have been reported with mean and standard deviation from runs on 5 differently seeded models. Please note that we shuffle the input order when testing, since norm-scaling uses running mean and average. That is, the order of in-distribution and OoD samples during testing is chosen at random which is a reasonable assumption for real-world deployment of such systems. This removes any bias that may be induced by the order of the inputs.

{\renewcommand{\arraystretch}{1.3}
\begin{table*}[!hbt]
\begin{adjustbox}{width=\textwidth}
\begin{tabular}{c|c|c|c|c|c|c}
\hline
Metric                 & In-Dist      & MSP  \cite{hendrycks17baseline}                   & ODIN  \cite{liang2020enhancing}                    & Mahalanobis    \cite{NIPS2018mahalanobis}              & Energy   \cite{liu2020energybased}             & Ours      \\ \hline
\multirow{4}{*}{AUROC $\uparrow$} & CIFAR-10     & 0.8234 $\pm$ 0.0538     & 0.8969 $\pm$ 0.0170       & 0.8533 $\pm$ 0.1269   & 0.8796 $\pm$ 0.0355   & \textbf { 0.9392 $\pm$ 0.0081 } \\ 
                       & CIFAR-100    & 0.6645 $\pm$ 0.1680     & 0.7307 $\pm$ 0.0903       & 0.6534 $\pm$ 0.2829   & 0.6721 $\pm$ 0.1706   & \textbf { 0.8931 $\pm$ 0.0027 } \\ 
                       & SVHN         & 0.9235 $\pm$ 0.0074     & 0.8359 $\pm$ 0.1671       & \textbf {0.9394 $\pm$ 0.0380}   & 0.9140 $\pm$ 0.0140   & 0.9181 $\pm$ 0.0064   \\  
                       & TinyImageNet & 0.5511 $\pm$ 0.0320     & 0.6420 $\pm$ 0.1335       & 0.7824 $\pm$ 0.1855   & 0.5529 $\pm$ 0.0971   & \textbf {0.7938 $\pm$ 0.0066}   \\ \cline{2-7}
                       & Average      & 0.7406 $\pm$ 0.1771	     & 0.7764 $\pm$ 0.1520       & 0.8071 $\pm$ 0.2097   & 0.7546 $\pm$ 0.1792   & \textbf{0.8860 $\pm$ 0.0643} \\\hline
\multirow{4}{*}{AUPR $\uparrow$}  & CIFAR-10     & 0.7803 $\pm$ 0.1113     & 0.8518 $\pm$ 0.0764       & 0.8261 $\pm$ 0.2116   & 0.8181 $\pm$ 0.0957   & \textbf { 0.9109 $\pm$ 0.0115 } \\ 
                       & CIFAR-100    & 0.6204 $\pm$ 0.1990     & 0.6563 $\pm$ 0.1766       & 0.6826 $\pm$ 0.2738   & 0.6143 $\pm$ 0.2022   & \textbf { 0.8397 $\pm$ 0.0036 } \\ 
                       & SVHN         & 0.7610 $\pm$ 0.1088     & 0.6933 $\pm$ 0.2162       & \textbf {0.8140 $\pm$ 0.1427}   & 0.7643 $\pm$ 0.1044   &  0.7760 $\pm$ 0.0183  \\  
                       & TinyImageNet & 0.5272 $\pm$ 0.1990     & 0.6144 $\pm$ 0.2203       & \textbf {0.7644 $\pm$ 0.2528}   & 0.5226 $\pm$ 0.1961   &  0.7457 $\pm$ 0.0068  \\ \cline{2-7}
                       & Average      & 0.6722 $\pm$ 0.1915     & 0.7040 $\pm$ 0.2028       & 0.7718 $\pm$ 0.2328   & 0.6798 $\pm$ 0.1966   & \textbf {0.8180 $\pm$ 0.0732} \\\hline
\multirow{4}{*}{FPR95 $\downarrow$} & CIFAR-10     & 0.6079 $\pm$ 0.1106     & 0.3167 $\pm$ 0.0857       & 0.4482 $\pm$ 0.3282   & 0.3222 $\pm$ 0.0562   & \textbf {  0.2160 $\pm$ 0.0236 }      \\  
                       & CIFAR-100    & 0.6577 $\pm$ 0.1304     & 0.6022 $\pm$ 0.1561       & 0.6757 $\pm$ 0.3510   & 0.6366 $\pm$ 0.1324   & \textbf { 0.3305 $\pm$ 0.0069 }     \\  
                       & SVHN         & 0.2660 $\pm$ 0.0442     & 0.5180 $\pm$ 0.2105       & \textbf {0.2164 $\pm$ 0.1037}   & 0.3603 $\pm$ 0.0724   & 0.3395 $\pm$ 0.0291     \\  
                       & TinyImageNet & 0.8261 $\pm$ 0.0658     & 0.7383 $\pm$ 0.2287       & 0.5355 $\pm$ 0.4109   & 0.8026 $\pm$ 0.0845   &  \textbf{0.5010 $\pm$ 0.0140}    \\ \cline{2-7}
                       & Average      & 0.5895 $\pm$ 0.2243     & 0.5438 $\pm$ 0.2355       & 0.4689 $\pm$ 0.3612   & 0.5304 $\pm$ 0.2184	  & \textbf {0.3578 $\pm$ 0.1019} \\\hline
\end{tabular}

\end{adjustbox}
\caption{Performance comparison of the proposed detection scheme i.e. MSP + Norm-scaling (averaged over 5 different seeds and OoD datasets) on various datasets (mean $\pm$ std) against OoD detection schemes. Expanded version of the table is available in the Appendix.}
\label{tab:comp_other}
\end{table*}}

\subsubsection{Norm-Scaling}
\label{sec:norm-scaling}
This subsection analyzes the effect of norm-scaling on OoD detection performance.  \cref{tab:norm_scaling_ood_perf} reports AUROC, AUPR and FPR95 of Maximum Softmax Probability (MSP) detector described by \cite{hendrycks17baseline} with and without norm-scaling. We choose  Maximum Softmax Probability (MSP) as baseline because this is the simplest OoD detector and is not influenced by other factors such as gradients (in case of ODIN, Mahalanobis) or noise and has been well studied in literature.

Norm-scaling achieves significant improvement across the board, the only exception being the AUROC performance on SVHN. The AUPR performance on SVHN is better than  un-scaled performance while norm-scaling does worse on AUROC performance on SVHN. We attribute this to the difference in the testset size, since AUROC numbers generally favor dominant class. On average, norm-scaling achieves 23\% improvement in AUROC, 24\% improvement in AUPR and a 31\% reduction in FPR95 over un-scaled  softmax scores (Baseline in \cref{tab:norm_scaling_ood_perf}).

\subsubsection{Multiple Thresholds}
An alternate solution to the issue of different angular similarly distributions for different classes is to threshold each class separately. This can be achieved by computing the performance metric separately for each class and then averaging the results. 

We implement this strategy in the following way. We infer over in-distribution and OoD samples, and obtain the predicted class for these inputs. We group the OoD score (in this case softmax score) and true label based on the predicted classes. For example, if the $i^{th}$ input is OoD and $j^{th}$ is in-dist and both are predicted as class 4, we put them in group 4. We do that for all the inputs. At the end we have K groups (one for each class in the training set). Each group contains OoD scores and corresponding true in-dist. vs OoD label. Using this we obtain the AUROC, AUPR and FPR95 metrics for each group and report the average result. If a group has no OoD samples, that group is skipped from the average computation. Further, we do this on 5 different seeds on 4 different in-distribution as described in \cref{sec:exp_setup}.

We compare the performance of using multiple thresholds on Maximum Softmax Probability (MSP) detector with norm-scaling and vanilla MSP in \cref{tab:norm_scaling_ood_perf}. Clearly, for all datasets other than SVHN we see a clear trend in performance, where performance of vanilla MSP is less than performance of MSP + Multi-Threshold which is intern less than the performance of MSP + Norm-scaling. From the results we infer that norm-scaling is better solution than using multiple thresholds.
We attribute the SVHN  performance outlier to the testset size of SVHN and to the equal weight assigned to each class' performance score. Equal weight for class' performance score ignores that some classes may get more OoD samples that others. However, weights based on group size are not also not preferable because this causes an imbalance for the in-distribution samples rather than OoD samples.

\subsubsection{Comparison}
\cref{tab:comp_other} compares the performance of the proposed norm-scaling method used with Maximum Softmax Probability (MSP) with ODIN \cite{liang2020enhancing}, Mahalanobis \cite{NIPS2018mahalanobis} energy based \cite{liu2020energybased} and MSP \cite{hendrycks17baseline} techniques. 
All the numbers reported in  \cref{tab:comp_other} are mean $\pm$ standard deviation obtained over 5 differently seeded and trained ResNet18 networks.  

Note, we have compared unsupervised methods. Unsupervised methods do not make any assumptions on the OoD datasets. Supervised methods such as \cite{hendrycks2019oe, liu2020energybased} use an auxiliary OoD dataset during training to improve OoD separability. This enforces a prior on the OoD dataset, but it is very difficult define such a prior \cite{liang2020enhancing}. This is because of the nature of OoD, the distribution prior is unknown. Furthermore, for \cite{liu2020energybased} there is a unsupervised and a supervised version; we compare with unsupervised one. 
On average, norm-scaling on MSP achieves \textbf{9.78\%} improvement in AUROC, \textbf{5.99\%} improvement in AUPR and \textbf{33.19\%} reduction in FPR95 over previous state-of-the-art methods.

\subsection{Temperature Scaling Perspective}

In this subsection we provide an alternate perspective/intuition for norm-scaling. We show that norm-scaling can be viewed as a parameter free version of temperature scaling. Temperature scaling is a technique \cite{platt1999probabilistic} that divides the logits by a temperature parameter $\tau$ (refer to \cref{eqn:temp_scaling}) and has been shown to improve OoD detection \cite{guo2017calibration, liang2020enhancing}.

To show norm-scaling as a parameter free version of temperature scaling we plot the expected calibration error (ECE) over various scaling temperatures (i.e. $\tau$ on the x-axis). We plot the ECE vs. temperature curves for norm-scaling and temperature scaling. Since norm-scaling described by \cref{eqn:norm_scaling} is parameter free, we modify it to introduce a temperature parameter $\tau$ resulting in the following equation

\begin{equation}
    z^{s}_{ij} = \frac{z_{ij} - \mu_j}{\tau \sigma_j}
\label{eqn:tau_norm_scaling}
\end{equation}
similarly we use temperature scaling given by
\begin{equation}
    z^{T}_{ij} = \frac{z_{ij}}{\tau}
\label{eqn:temp_scaling}
\end{equation}

\begin{figure}[hbt!]
    \centering
    \includegraphics[scale=0.27]{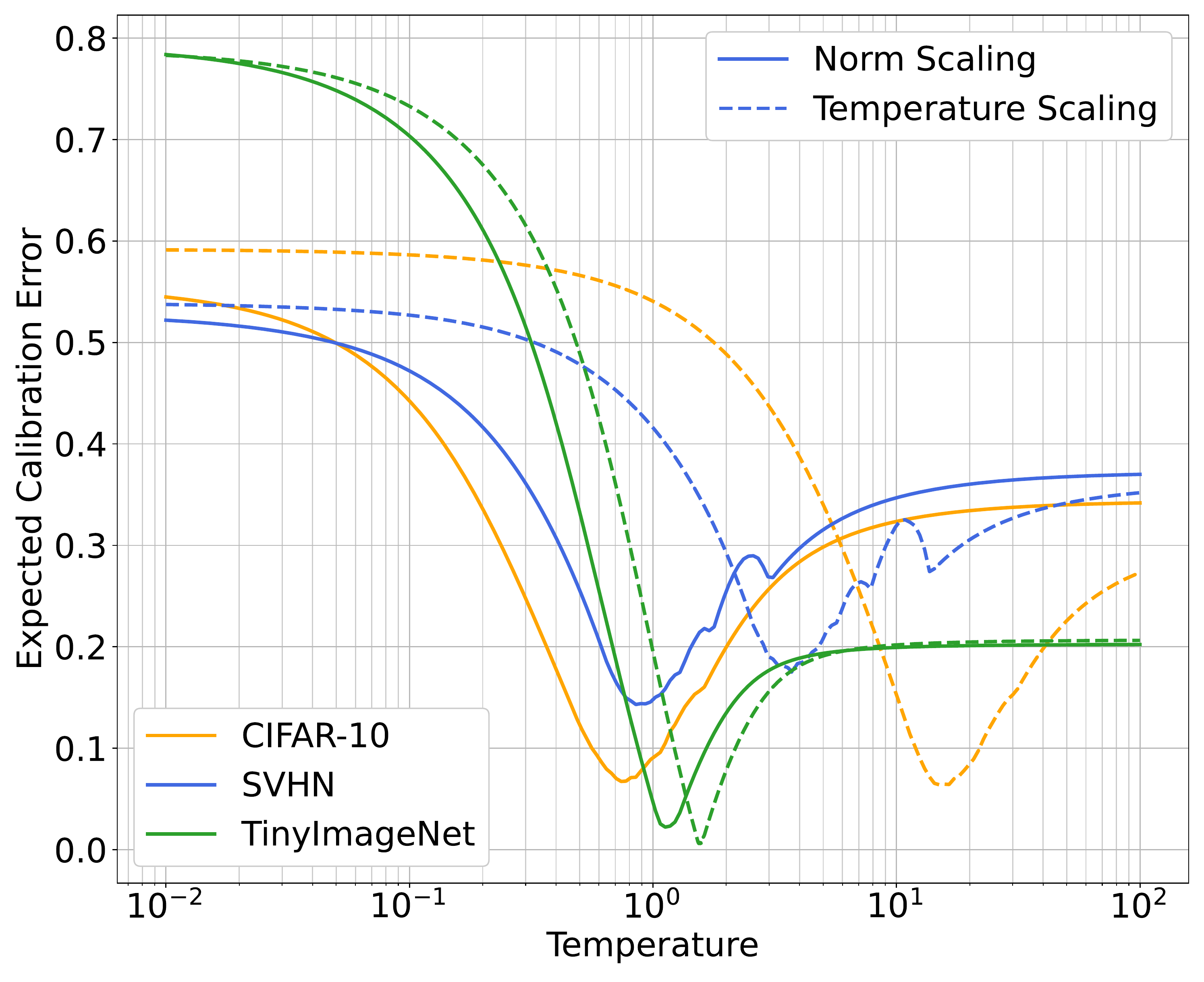}
    \caption{Norm-scaling's minima for expected calibration error lies close to a temperature of 1 (all the solid lines have a minima close to 1) for different datasets, while for temperature scaling the minima temperature varies by several orders of magnitude for different dataset.}
    \label{fig:temp_vs_norm}
\end{figure}

\cref{fig:temp_vs_norm} shows ECE vs. temperature for norm-scaling and temperature-scaling on a ResNet-18 network trained on various datasets. We observe that the optimal temperature $\tau^{norm}_{opt}$ (given by the lowest ECE value) is very close to 1 for norm-scaling across different datasets, while the optimal temperature $\tau^{temp}_{opt}$ for temperature scaling varies by orders of magnitude depending on the training dataset. Therefore, by setting $\tau^{norm}_{opt} = 1$ in \cref{eqn:tau_norm_scaling} one can achieve near optimal calibration. Setting $\tau^{norm}_{opt} = 1$ in \cref{eqn:tau_norm_scaling} results in the norm-scaling as proposed in \cref{sec:norm-scaling}.
Therefore, norm-scaling can be viewed as a parameter free version of temperature scaling.

\section{Discussion and Conclusion}
In this paper we observe that different classes have different distributions of angular similarity. Angular similarity encodes uncertainty, thus a single similarity (or similarity proxy) value represents different uncertainties for different classes. This implies that a single threshold across various classes is not suited to separate in-distribution data from OoD data. To address this issue we propose norm-scaling, a technique that normalizes the logits of the neural network to ensure consistent uncertainty value between different classes. We show that norm scaling significantly improves OoD performance and is better than the using a separate threshold for each class.
Further, we also provide a temperature scaling perspective of the proposed norm-scaling approach showing that norm-scaling can be interpreted as a parameter free version of temperature scaling.
Our experimental results show that the proposed norm-scaling when used with maximum softmax probability (MSP) \cite{hendrycks17baseline} detector achieves on average (across various datasets and seeds) \textbf{9.78\%} improvement in AUROC, \textbf{5.99\%} improvement in AUPR and \textbf{33.19\%} reduction in FPR95 over previous state-of-the-art methods.

\section{Acknowledgement}
This work was supported in part by the Center for Brain Inspired Computing (C-BRIC), one of the six centers in JUMP, a Semiconductor Research Corporation (SRC) program sponsored by DARPA, by the Semiconductor Research Corporation, the National Science Foundation, Intel Corporation, the DoD Vannevar Bush Fellowship, and by the U.S. Army Research Laboratory and the U.K. Ministry of Defence under Agreement Number W911NF-16-3-0001.

{\small
\bibliographystyle{ieee_fullname}
\bibliography{egbib}
}

\section*{Appendix}
\subsection*{Code and Pretrained Models}
Anonymized versions of the code an pretrained models are available at \url{https://anonymous.4open.science/r/Norm-Scaling-08D9/}

\subsection*{Expanded Results}
This section contains the expanded results for the OoD performance of various techniques and the proposed technique (see following pages).
{\renewcommand{\arraystretch}{1.5}
\begin{table*}[htb!]
\begin{adjustbox}{width=\textwidth}
\begin{tabular}{c|c|ccc|ccc|ccc}
\hline
\multicolumn{11}{c}{Expanded results for Maximum Softmax Probability (MSP) detector, MSP with multiple thresholds and MSP with norm-scaling.}                                                                                                               \\ \hline
\multirow{2}{*}{In-Dataset}          &  \multirow{2}{*}{OoD Dataset}            & \multicolumn{3}{c}{AUROC}           & \multicolumn{3}{c}{AUPR}            & \multicolumn{3}{c}{FPR95}           \\ \cline{3-11}
 &                                                  & Baseline                 & Multi-Thresh.               & Norm-Scaling              & Baseline                 & Multi-Thresh.               & Norm-Scaling                &   Baseline                 & Multi-Thresh.               & Norm-Scaling              \\ \hline
\multirow{8}{*}{CIFAR-10}     & Gaussian Noise      & 0.6991 $\pm$ 0.2232 & 0.7934 $\pm$ 0.1970 & 0.9882 $\pm$ 0.0094 & 0.6451 $\pm$ 0.2250 & 0.7724 $\pm$ 0.1751 &	0.9803 $\pm$ 0.0186 & 0.6103 $\pm$ 0.3818 & 0.5283 $\pm$ 0.2664 & 0.0359 $\pm$ 0.0229\\
                              & Uniform Noise       & 0.8722 $\pm$ 0.0308 & 0.9422 $\pm$ 0.0103 & 0.9864 $\pm$ 0.0077 & 0.8005 $\pm$ 0.0413 & 0.9315 $\pm$ 0.0209 &	0.9774 $\pm$ 0.0138 & 0.3513 $\pm$ 0.2029 & 0.4201 $\pm$ 0.1041	& 0.0445 $\pm$ 0.0231\\
                              & SVHN                & 0.8111 $\pm$ 0.0901 & 0.8997 $\pm$ 0.0135 & 0.9522 $\pm$ 0.0120 & 0.9005 $\pm$ 0.0402 & 0.8715 $\pm$ 0.0053 & 0.9740 $\pm$ 0.0070 & 0.6102 $\pm$ 0.2288 & 0.4573 $\pm$ 0.0286	& 0.1610 $\pm$ 0.0349\\
                              & Textures            & 0.8448 $\pm$ 0.0089 & 0.8643 $\pm$ 0.0067 & 0.9087 $\pm$ 0.0092 & 0.7311 $\pm$ 0.0058 & 0.9204 $\pm$ 0.0014 & 0.8179 $\pm$ 0.0146 & 0.6698 $\pm$ 0.1196 & 0.5883 $\pm$ 0.0088	& 0.3054 $\pm$ 0.0291\\
                              & LSUN                & 0.8598 $\pm$ 0.0177 & 0.8893 $\pm$ 0.0111 & 0.9312 $\pm$ 0.0058 & 0.6262 $\pm$ 0.0203 & 0.9618 $\pm$ 0.0044 & 0.7818 $\pm$ 0.0166 & 0.6217 $\pm$ 0.1507 & 0.5420 $\pm$ 0.0213	& 0.2522 $\pm$ 0.0129\\
                              & TinyImageNet        & 0.8344 $\pm$ 0.0143 & 0.8319 $\pm$ 0.0148 & 0.8923 $\pm$ 0.0063 & 0.8169 $\pm$ 0.0078 & 0.8592 $\pm$ 0.0064 & 0.8747 $\pm$ 0.0073 & 0.7070 $\pm$ 0.1248 & 0.6453 $\pm$ 0.0152	& 0.3908 $\pm$ 0.0223\\
                              & Places365           & 0.8422 $\pm$ 0.0155 & 0.8559 $\pm$ 0.0122 & 0.9156 $\pm$ 0.0064 & 0.9418 $\pm$ 0.0040 & 0.7459 $\pm$ 0.0066 & 0.9702 $\pm$ 0.0024 & 0.6851 $\pm$ 0.1368 & 0.6083 $\pm$ 0.0155	& 0.3220 $\pm$ 0.0199\\ \cline{2-11}
                              & Average             & 0.8234 $\pm$ 0.0538 & 0.8681 $\pm$ 0.0379 & 0.9392 $\pm$ 0.0081 & 0.7803 $\pm$ 0.1113 & 0.8661 $\pm$ 0.0314 & 0.9109 $\pm$ 0.0115 & 0.6079 $\pm$ 0.1106 & 0.5414 $\pm$ 0.0657	& 0.2160 $\pm$ 0.0236\\ \hline

\multirow{8}{*}{CIFAR-100}    & Gaussian Noise      & 0.2596 $\pm$ 0.0665 & 0.4333 $\pm$ 0.0840	& 1.0000 $\pm$ 0.0000 & 0.3647 $\pm$ 0.0181 & 0.2978 $\pm$ 0.0979 & 1.0000 $\pm$ 0.0000 & 0.8237 $\pm$ 0.0798 & 1.0000 $\pm$ 0.0000	& 0.0000 $\pm$ 0.0000\\
                              & Uniform Noise       & 0.7941 $\pm$ 0.0627 & 0.9046 $\pm$ 0.0247	& 1.0000 $\pm$ 0.0000 & 0.6889 $\pm$ 0.0805 & 0.8423 $\pm$ 0.0509 & 1.0000 $\pm$ 0.0000 & 0.3759 $\pm$ 0.0618 & 0.5220 $\pm$ 0.1277	& 0.0000 $\pm$ 0.0001\\
                              & SVHN                & 0.7426 $\pm$ 0.0273 & 0.8383 $\pm$ 0.0058	& 0.9701 $\pm$ 0.0015 & 0.8507 $\pm$ 0.0173 & 0.8618 $\pm$ 0.0128 & 0.9875 $\pm$ 0.0008 & 0.6120 $\pm$ 0.0456 & 0.6866 $\pm$ 0.0291	& 0.1534 $\pm$ 0.0067\\
                              & Textures            & 0.6948 $\pm$ 0.0038 & 0.7772 $\pm$ 0.0033	& 0.8201 $\pm$ 0.0054 & 0.5075 $\pm$ 0.0046 & 0.8993 $\pm$ 0.0011 & 0.6665 $\pm$ 0.0077 & 0.7322 $\pm$ 0.0079 & 0.7939 $\pm$ 0.0072	& 0.5110 $\pm$ 0.0129\\
                              & LSUN                & 0.7020 $\pm$ 0.0051 & 0.8011 $\pm$ 0.0062	& 0.8255 $\pm$ 0.0059 & 0.3591 $\pm$ 0.0061 & 0.9351 $\pm$ 0.0018 & 0.5350 $\pm$ 0.0112 & 0.7094 $\pm$ 0.0059 & 0.7669 $\pm$ 0.0262	& 0.5405 $\pm$ 0.0150\\
                              & TinyImageNet        & 0.7382 $\pm$ 0.0026 & 0.7473 $\pm$ 0.0034	& 0.7880 $\pm$ 0.0029 & 0.6900 $\pm$ 0.0042 & 0.7758 $\pm$ 0.0013 & 0.7474 $\pm$ 0.0043 & 0.6518 $\pm$ 0.0025 & 0.8227 $\pm$ 0.0080	& 0.6113 $\pm$ 0.0029\\
                              & Places365           & 0.7201 $\pm$ 0.0040 & 0.7777 $\pm$ 0.0032	& 0.8481 $\pm$ 0.0035 & 0.8818 $\pm$ 0.0022 & 0.6939 $\pm$ 0.0026 & 0.9420 $\pm$ 0.0015 & 0.6992 $\pm$ 0.0066 & 0.7861 $\pm$ 0.0052	& 0.4975 $\pm$ 0.0107\\ \cline{2-11}
                              & Average             & 0.6645 $\pm$ 0.1680 & 0.7542 $\pm$ 0.0187	& 0.8931 $\pm$ 0.0027 & 0.6204 $\pm$ 0.1990 & 0.7580 $\pm$ 0.0241 & 0.8397 $\pm$ 0.0036 & 0.6577 $\pm$ 0.1304 & 0.7683 $\pm$ 0.0290 & 0.3305 $\pm$ 0.0069\\ \hline
          
\multirow{8}{*}{SVHN}         & Gaussian Noise      & 0.9218 $\pm$ 0.0141 & 0.9532 $\pm$ 0.0041	& 0.9327 $\pm$ 0.0103 & 0.7770 $\pm$ 0.0298 & 0.9806 $\pm$ 0.0028 & 0.8245 $\pm$ 0.0303 & 0.2575 $\pm$ 0.0560 & 0.3000 $\pm$ 0.0259	& 0.2617 $\pm$ 0.0268\\
                              & Uniform Noise       & 0.9247 $\pm$ 0.0061 & 0.9435 $\pm$ 0.0065	& 0.9204 $\pm$ 0.0086 & 0.7845 $\pm$ 0.0061 & 0.9793 $\pm$ 0.0021 & 0.7943 $\pm$ 0.0263 & 0.2494 $\pm$ 0.0335 & 0.3615 $\pm$ 0.0234	& 0.3113 $\pm$ 0.0237\\
                              & CIFAR-100           & 0.9267 $\pm$ 0.0042 & 0.9323 $\pm$ 0.0034	& 0.9137 $\pm$ 0.0050 & 0.7993 $\pm$ 0.0067 & 0.9737 $\pm$ 0.0015 & 0.8018 $\pm$ 0.0093 & 0.2552 $\pm$ 0.0257 & 0.4126 $\pm$ 0.0168	& 0.3767 $\pm$ 0.0337\\
                              & Textures            & 0.9066 $\pm$ 0.0059 & 0.9266 $\pm$ 0.0026	& 0.8917 $\pm$ 0.0054 & 0.6651 $\pm$ 0.0154 & 0.9785 $\pm$ 0.0014 & 0.6602 $\pm$ 0.0193 & 0.3723 $\pm$ 0.0364 & 0.4136 $\pm$ 0.0122	& 0.4515 $\pm$ 0.0346\\
                              & LSUN                & 0.9258 $\pm$ 0.0051 & 0.9439 $\pm$ 0.0010	& 0.9206 $\pm$ 0.0050 & 0.5614 $\pm$ 0.0133	& 0.9926 $\pm$ 0.0004 & 0.5987 $\pm$ 0.0257 & 0.2534 $\pm$ 0.0260 & 0.3658 $\pm$ 0.0076	& 0.3335 $\pm$ 0.0313\\
                              & TinyImageNet        & 0.9308 $\pm$ 0.0037 & 0.9390 $\pm$ 0.0025	& 0.9230 $\pm$ 0.0043 & 0.8071 $\pm$ 0.0066 & 0.9771 $\pm$ 0.0012 & 0.8169 $\pm$ 0.0109 & 0.2306 $\pm$ 0.0221 & 0.3858 $\pm$ 0.0090	& 0.3269 $\pm$ 0.0234\\
                              & Places365           & 0.9281 $\pm$ 0.0047 & 0.9400 $\pm$ 0.0019	& 0.9242 $\pm$ 0.0065 & 0.9324 $\pm$ 0.0034 & 0.9360 $\pm$ 0.0042 & 0.9358 $\pm$ 0.0063 & 0.2436 $\pm$ 0.0256 & 0.3824 $\pm$ 0.0085	& 0.3146 $\pm$ 0.0299\\ \cline{2-11}
                              & Average             & 0.9235 $\pm$ 0.0074 & 0.9398 $\pm$ 0.0031	& 0.9181 $\pm$ 0.0064 & 0.7610 $\pm$ 0.1088 & 0.9740 $\pm$ 0.0019 & 0.7760 $\pm$ 0.0183 & 0.2660 $\pm$ 0.0442 & 0.3745 $\pm$ 0.0148	& 0.3395 $\pm$ 0.0291\\ \hline
                              
\multirow{8}{*}{TinyImageNet} & Gaussian Noise      & 0.5506 $\pm$ 0.1104 & 0.7168 $\pm$ 0.0614	& 0.9967 $\pm$ 0.0020 & 0.4911 $\pm$ 0.0665 & 0.7179 $\pm$ 0.0666 &	0.9966 $\pm$ 0.0021 & 0.7374 $\pm$ 0.1040 & 0.9286 $\pm$ 0.0342	& 0.0149 $\pm$ 0.0103 \\
                              & Uniform Noise       & 0.2977 $\pm$ 0.0686 & 0.6399 $\pm$ 0.0424	& 0.9936 $\pm$ 0.0025 & 0.3726 $\pm$ 0.0220 & 0.7938 $\pm$ 0.0253 &	0.9935 $\pm$ 0.0026 & 0.9064 $\pm$ 0.0255 & 0.9398 $\pm$ 0.0348	& 0.0317 $\pm$ 0.0139 \\
                              & SVHN                & 0.6747 $\pm$ 0.0258 & 0.7319 $\pm$ 0.0211	& 0.9347 $\pm$ 0.0098 & 0.7956 $\pm$ 0.0237 & 0.7719 $\pm$ 0.0072 &	0.9720 $\pm$ 0.0047 & 0.7187 $\pm$ 0.0201 & 0.8395 $\pm$ 0.0516	& 0.3118 $\pm$ 0.0335 \\
                              & CIFAR-100           & 0.5774 $\pm$ 0.0034 & 0.5989 $\pm$ 0.0029	& 0.5999 $\pm$ 0.0035 & 0.5491 $\pm$ 0.0038 & 0.6356 $\pm$ 0.0022 &	0.5675 $\pm$ 0.0031 & 0.8718 $\pm$ 0.0035 & 0.9121 $\pm$ 0.0079	& 0.8534 $\pm$ 0.0040 \\
                              & Textures            & 0.5555 $\pm$ 0.0078 & 0.6245 $\pm$ 0.0068	& 0.6580 $\pm$ 0.0163 & 0.3801 $\pm$ 0.0099 & 0.8090 $\pm$ 0.0030 & 0.4754 $\pm$ 0.0272 & 0.8702 $\pm$ 0.0056 & 0.9146 $\pm$ 0.0161	& 0.7853 $\pm$ 0.0124  \\
                              & LSUN                & 0.5897 $\pm$ 0.0051 & 0.6692 $\pm$ 0.0107	& 0.6717 $\pm$ 0.0063 & 0.2709 $\pm$ 0.0041 & 0.8693 $\pm$ 0.0042 &	0.3382 $\pm$ 0.0055 & 0.8451 $\pm$ 0.0044 & 0.8612 $\pm$ 0.0191	& 0.7626 $\pm$ 0.0130 \\
                              & Places365           & 0.6117 $\pm$ 0.0045 & 0.6479 $\pm$ 0.0030	& 0.7022 $\pm$ 0.0061 & 0.8309 $\pm$ 0.0014 & 0.4983 $\pm$ 0.0027 &	0.8764 $\pm$ 0.0023 & 0.8334 $\pm$ 0.0042 & 0.8845 $\pm$ 0.0069	& 0.7472 $\pm$ 0.0109 \\ \cline{2-11}
                              & Average             & 0.5510 $\pm$ 0.1104 & 0.6613 $\pm$ 0.0212	& 0.7938 $\pm$ 0.0066 & 0.5272 $\pm$ 0.1990 & 0.7280 $\pm$ 0.0159 & 0.7457 $\pm$ 0.0068 & 0.8261 $\pm$ 0.0658 & 0.8972 $\pm$ 0.0244	& 0.5010 $\pm$ 0.0140 \\ \hline
\end{tabular}
\end{adjustbox}
\vspace{2mm}
\caption{Expanded results for Maximum Softmax Probability (MSP i.e. ``Baseline'') detector, MSP with multiple thresholds and MSP with norm-scaling.}
\label{tab:exp_msp}
\end{table*}}

{\renewcommand{\arraystretch}{1.5}
\begin{table*}[htb!]
\begin{adjustbox}{width=\textwidth}
\begin{tabular}{c|c|ccc|ccc|ccc}
\hline
\multicolumn{11}{c}{Expanded results for ODIN, Energy and Mahalanobis distance based methods }                                                                                                               \\ \hline
\multirow{2}{*}{In-Dataset}          &  \multirow{2}{*}{OoD Dataset}            & \multicolumn{3}{c}{AUROC}           & \multicolumn{3}{c}{AUPR}            & \multicolumn{3}{c}{FPR95}           \\ \cline{3-11}
 &                                                  & ODIN                 & Energy               & Mahalanobis              & ODIN                 & Energy               & Mahalanobis                & ODIN                 & Energy               & Mahalanobis            \\ \hline
\multirow{8}{*}{CIFAR-10} 
 & Gaussian Noise   & 0.9059 $\pm$ 0.0866 & 0.7997 $\pm$ 0.1342 & 0.9998 $\pm$ 0.0003   & 0.8378 $\pm$ 0.1369 & 0.6885 $\pm$ 0.1384 & 0.9996 $\pm$ 0.0006   & 0.1691 $\pm$ 0.1213 & 0.2975 $\pm$ 0.1625 & 0.0005 $\pm$ 0.0007  \\ 
 & Uniform Noise    & 0.8624 $\pm$ 0.0585 & 0.8719 $\pm$ 0.0356 & 0.9969 $\pm$ 0.0032   & 0.7709 $\pm$ 0.0863 & 0.7576 $\pm$ 0.0457 & 0.9904 $\pm$ 0.0084   & 0.2807 $\pm$ 0.1075 & 0.2446 $\pm$ 0.0688 & 0.0069 $\pm$ 0.0083  \\ 
 & SVHN             & 0.9127 $\pm$ 0.0441 & 0.8880 $\pm$ 0.0354 & 0.9552 $\pm$ 0.0185   & 0.9490 $\pm$ 0.0260 & 0.9268 $\pm$ 0.0191 & 0.9808 $\pm$ 0.0079   & 0.2489 $\pm$ 0.1176 & 0.2773 $\pm$ 0.0891 & 0.2338 $\pm$ 0.1159  \\ 
 & Textures         & 0.8993 $\pm$ 0.0202 & 0.8849 $\pm$ 0.0114 & 0.8830 $\pm$ 0.0525   & 0.8237 $\pm$ 0.0266 & 0.7669 $\pm$ 0.0132 & 0.8729 $\pm$ 0.0457   & 0.3650 $\pm$ 0.0836 & 0.3722 $\pm$ 0.0541 & 0.6542 $\pm$ 0.2605  \\ 
 & LSUN             & 0.9151 $\pm$ 0.0049 & 0.9201 $\pm$ 0.0041 & 0.7151 $\pm$ 0.0210   & 0.7474 $\pm$ 0.0074 & 0.7494 $\pm$ 0.0104 & 0.3715 $\pm$ 0.0225   & 0.3184 $\pm$ 0.0258 & 0.2860 $\pm$ 0.0119 & 0.7212 $\pm$ 0.0227  \\ 
 & TinyImageNet     & 0.8840 $\pm$ 0.0130 & 0.8899 $\pm$ 0.0072 & 0.7052 $\pm$ 0.0187   & 0.8696 $\pm$ 0.0123 & 0.8724 $\pm$ 0.0071 & 0.6839 $\pm$ 0.0174   & 0.4406 $\pm$ 0.0560 & 0.4086 $\pm$ 0.0326 & 0.7819 $\pm$ 0.0402  \\ 
 & Places365        & 0.8987 $\pm$ 0.0081 & 0.9030 $\pm$ 0.0059 & 0.7180 $\pm$ 0.0133   & 0.9642 $\pm$ 0.0028 & 0.9650 $\pm$ 0.0021 & 0.8837 $\pm$ 0.0082   & 0.3942 $\pm$ 0.0390 & 0.3693 $\pm$ 0.0225 & 0.7386 $\pm$ 0.0134  \\ \cline{2-11}
 & Average          & 0.8969 $\pm$ 0.0170 & 0.8796 $\pm$ 0.0355 & 0.8533 $\pm$ 0.1269   & 0.8518 $\pm$ 0.0764 & 0.8181 $\pm$ 0.0957 & 0.8261 $\pm$ 0.2116   & 0.3167 $\pm$ 0.0857 & 0.3222 $\pm$ 0.0562 & 0.4482 $\pm$ 0.3282  \\ \hline

\multirow{8}{*}{CIFAR-100}     
& Gaussian Noise & 0.5267 $\pm$ 0.0759 & 0.2613 $\pm$ 0.0566 & 0.1173 $\pm$ 0.1434   & 0.4631 $\pm$ 0.0373 & 0.3657 $\pm$ 0.0169 & 0.3912 $\pm$ 0.1109   & 0.5671 $\pm$ 0.0788 & 0.8050 $\pm$ 0.0531 & 1.0000 $\pm$ 0.0000  \\ 
& Uniform Noise  & 0.8447 $\pm$ 0.0539 & 0.7004 $\pm$ 0.0573 & 0.9974 $\pm$ 0.0024   & 0.7173 $\pm$ 0.0776 & 0.5664 $\pm$ 0.0464 & 0.9916 $\pm$ 0.0061   & 0.2490 $\pm$ 0.0735 & 0.3918 $\pm$ 0.0625 & 0.0054 $\pm$ 0.0058  \\ 
& SVHN           & 0.7614 $\pm$ 0.0249 & 0.7763 $\pm$ 0.0206 & 0.9351 $\pm$ 0.0158   & 0.8480 $\pm$ 0.0182 & 0.8414 $\pm$ 0.0158 & 0.9667 $\pm$ 0.0072   & 0.5813 $\pm$ 0.0369 & 0.5026 $\pm$ 0.0291 & 0.2889 $\pm$ 0.0796  \\ 
& Textures       & 0.7347 $\pm$ 0.0060 & 0.7045 $\pm$ 0.0072 & 0.8641 $\pm$ 0.0245   & 0.5533 $\pm$ 0.0068 & 0.5062 $\pm$ 0.0065 & 0.8429 $\pm$ 0.0223   & 0.7023 $\pm$ 0.0105 & 0.7266 $\pm$ 0.0132 & 0.7253 $\pm$ 0.1058  \\ 
& LSUN           & 0.7323 $\pm$ 0.0069 & 0.7195 $\pm$ 0.0057 & 0.5407 $\pm$ 0.0174   & 0.3957 $\pm$ 0.0093 & 0.3714 $\pm$ 0.0073 & 0.2365 $\pm$ 0.0084   & 0.7386 $\pm$ 0.0127 & 0.7125 $\pm$ 0.0101 & 0.8965 $\pm$ 0.0132  \\ 
& TinyImageNet   & 0.7635 $\pm$ 0.0033 & 0.7855 $\pm$ 0.0024 & 0.5478 $\pm$ 0.0115   & 0.7197 $\pm$ 0.0051 & 0.7478 $\pm$ 0.0033 & 0.5350 $\pm$ 0.0087   & 0.6707 $\pm$ 0.0064 & 0.6305 $\pm$ 0.0018 & 0.9189 $\pm$ 0.0047  \\ 
& Places365      & 0.7515 $\pm$ 0.0062 & 0.7569 $\pm$ 0.0040 & 0.5711 $\pm$ 0.0193   & 0.8972 $\pm$ 0.0032 & 0.9009 $\pm$ 0.0020 & 0.8140 $\pm$ 0.0098   & 0.7064 $\pm$ 0.0097 & 0.6873 $\pm$ 0.0055 & 0.8949 $\pm$ 0.0121  \\ \cline{2-11}
& Average        & 0.7307 $\pm$ 0.0903 & 0.6721 $\pm$ 0.1706 & 0.6534 $\pm$ 0.2829   & 0.6563 $\pm$ 0.1766 & 0.6143 $\pm$ 0.2022 & 0.6826 $\pm$ 0.2738   & 0.6022 $\pm$ 0.1561 & 0.6366 $\pm$ 0.1324 & 0.6757 $\pm$ 0.3510  \\ \hline
          
\multirow{8}{*}{SVHN}         
& Gaussian Noise & 0.8929 $\pm$ 0.0257 & 0.9150 $\pm$ 0.0220 & 0.9996 $\pm$ 0.0003   & 0.7453 $\pm$ 0.0509 & 0.7778 $\pm$ 0.0484 & 0.9950 $\pm$ 0.0041   & 0.4429 $\pm$ 0.1016 & 0.3248 $\pm$ 0.0859 & 0.0006 $\pm$ 0.0003  \\ 
& Uniform Noise  & 0.4328 $\pm$ 0.0662 & 0.9007 $\pm$ 0.0148 & 0.8598 $\pm$ 0.0784   & 0.2320 $\pm$ 0.0232 & 0.7371 $\pm$ 0.0478 & 0.6224 $\pm$ 0.1267   & 0.9430 $\pm$ 0.0224 & 0.3753 $\pm$ 0.0504 & 0.3829 $\pm$ 0.1456	\\ 
& CIFAR-100      & 0.9216 $\pm$ 0.0105 & 0.9229 $\pm$ 0.0072 & 0.9456 $\pm$ 0.0041   & 0.8360 $\pm$ 0.0145 & 0.8174 $\pm$ 0.0136 & 0.8574 $\pm$ 0.0131   & 0.3803 $\pm$ 0.0717 & 0.3321 $\pm$ 0.0450 & 0.2148 $\pm$ 0.0207  \\ 
& Textures       & 0.8375 $\pm$ 0.0083 & 0.8862 $\pm$ 0.0080 & 0.9502 $\pm$ 0.0033   & 0.6088 $\pm$ 0.0133 & 0.6555 $\pm$ 0.0267 & 0.8502 $\pm$ 0.0081   & 0.7294 $\pm$ 0.0526 & 0.5292 $\pm$ 0.0311 & 0.2374 $\pm$ 0.0233  \\ 
& LSUN           & 0.9216 $\pm$ 0.0122 & 0.9203 $\pm$ 0.0100 & 0.9399 $\pm$ 0.0045   & 0.6500 $\pm$ 0.0331 & 0.5992 $\pm$ 0.0367 & 0.5848 $\pm$ 0.0374   & 0.3831 $\pm$ 0.0715 & 0.3448 $\pm$ 0.0522 & 0.2153 $\pm$ 0.0141  \\ 
& TinyImageNet   & 0.9251 $\pm$ 0.0091 & 0.9283 $\pm$ 0.0067 & 0.9424 $\pm$ 0.0035   & 0.8422 $\pm$ 0.0142 & 0.8266 $\pm$ 0.0155 & 0.8469 $\pm$ 0.0150   & 0.3628 $\pm$ 0.0570 & 0.2979 $\pm$ 0.0351 & 0.2247 $\pm$ 0.0143  \\ 
& Places365      & 0.9201 $\pm$ 0.0111 & 0.9243 $\pm$ 0.0086 & 0.9386 $\pm$ 0.0041   & 0.9386 $\pm$ 0.0076 & 0.9363 $\pm$ 0.0076 & 0.9414 $\pm$ 0.0065   & 0.3843 $\pm$ 0.0656 & 0.3179 $\pm$ 0.0443 & 0.2389 $\pm$ 0.0150  \\ \cline{2-11}
& Average        & 0.8359 $\pm$ 0.1671 & 0.9140 $\pm$ 0.0140 & 0.9394 $\pm$ 0.0380   & 0.6933 $\pm$ 0.2162 & 0.7643 $\pm$ 0.1044 & 0.8140 $\pm$ 0.1427   & 0.5180 $\pm$ 0.2105 & 0.3603 $\pm$ 0.0724 & 0.2164 $\pm$ 0.1037  \\ \hline

\multirow{8}{*}{TinyImageNet} 
& Gaussian Noise    & 0.9353 $\pm$ 0.0691 & 0.5227 $\pm$ 0.1378 & 0.9970 $\pm$ 0.0052   & 0.9130 $\pm$ 0.0960 & 0.4717 $\pm$ 0.0648 & 0.9978 $\pm$ 0.0029   & 0.1958 $\pm$ 0.1672 & 0.6961 $\pm$ 0.1173 & 0.0013 $\pm$ 0.0017  \\ 
& Uniform Noise     & 0.4737 $\pm$ 0.0700 & 0.3399 $\pm$ 0.0730 & 0.9932 $\pm$ 0.0039   & 0.4421 $\pm$ 0.0348 & 0.3868 $\pm$ 0.0259 & 0.9764 $\pm$ 0.0131   & 0.7836 $\pm$ 0.0461 & 0.8750 $\pm$ 0.0345 & 0.0121 $\pm$ 0.0065  \\ 
& SVHN              & 0.6754 $\pm$ 0.0380 & 0.6734 $\pm$ 0.0499 & 0.9486 $\pm$ 0.0147   & 0.8018 $\pm$ 0.0315 & 0.7781 $\pm$ 0.0410 & 0.9684 $\pm$ 0.0116   & 0.7237 $\pm$ 0.0281 & 0.6508 $\pm$ 0.0418 & 0.1831 $\pm$ 0.0477  \\ 
& CIFAR-100         & 0.6016 $\pm$ 0.0034 & 0.5778 $\pm$ 0.0041 & 0.6054 $\pm$ 0.0121   & 0.5805 $\pm$ 0.0037 & 0.5500 $\pm$ 0.0031 & 0.5790 $\pm$ 0.0097   & 0.8754 $\pm$ 0.0051 & 0.8775 $\pm$ 0.0042 & 0.9083 $\pm$ 0.0206  \\ 
& Textures          & 0.5576 $\pm$ 0.0067 & 0.5620 $\pm$ 0.0187 & 0.7947 $\pm$ 0.0349   & 0.4226 $\pm$ 0.0084 & 0.3813 $\pm$ 0.0185 & 0.7730 $\pm$ 0.0208   & 0.9145 $\pm$ 0.0032 & 0.8539 $\pm$ 0.0055 & 0.8978 $\pm$ 0.1083  \\ 
& LSUN              & 0.6252 $\pm$ 0.0052 & 0.5810 $\pm$ 0.0081 & 0.5613 $\pm$ 0.0339   & 0.3022 $\pm$ 0.0056 & 0.2612 $\pm$ 0.0063 & 0.2450 $\pm$ 0.0181   & 0.8381 $\pm$ 0.0056 & 0.8436 $\pm$ 0.0051 & 0.8685 $\pm$ 0.0286  \\ 
& Places365         & 0.6251 $\pm$ 0.0033 & 0.6135 $\pm$ 0.0038 & 0.5765 $\pm$ 0.0292   & 0.8388 $\pm$ 0.0020 & 0.8287 $\pm$ 0.0019 & 0.8112 $\pm$ 0.0161   & 0.8373 $\pm$ 0.0046 & 0.8212 $\pm$ 0.0038 & 0.8776 $\pm$ 0.0211  \\ \cline{2-11}
& Average           & 0.6420 $\pm$ 0.1335 & 0.5529 $\pm$ 0.0971 & 0.7824 $\pm$ 0.1855   & 0.6144 $\pm$ 0.2203 & 0.5226 $\pm$ 0.1961 & 0.7644 $\pm$ 0.2528   & 0.7383 $\pm$ 0.2287 & 0.8026 $\pm$ 0.0845 & 0.5355 $\pm$ 0.4109  \\ \hline
\multicolumn{2}{c}{Average}  
&  0.7764 $\pm$ 0.1520 & 0.7546 $\pm$ 0.1792 & 0.8071 $\pm$ 0.2097   & 0.7040 $\pm$ 0.2028 & 0.6798 $\pm$ 0.1966 & 0.7718 $\pm$ 0.2328   & 0.5438 $\pm$ 0.2355 & 0.5304 $\pm$ 0.2184 & 0.4689 $\pm$ 0.3612  \\  \hline
\end{tabular}
\end{adjustbox}
\vspace{2mm}
\caption{Expanded results for ODIN, Energy and Mahalanobis distance based methods.}
\label{tab:exp_odin}
\end{table*}}

{\renewcommand{\arraystretch}{1.2}
\begin{table*}[htb!]
\centering
\begin{adjustbox}{width=0.8\textwidth}
\begin{tabular}{c|c|c|c|c}
\hline
\multicolumn{5}{c}{Expanded results for the proposed method (i.e. MSP + norm-scaling)}                                                                                                               \\ \hline
\multirow{1}{*}{In-Dataset}          &  \multirow{1}{*}{OoD Dataset}            & \multicolumn{1}{c}{AUROC}           & \multicolumn{1}{c}{AUPR}            & \multicolumn{1}{c}{FPR95}           \\ \hline
\multirow{5}{*}{CIFAR-10} 
 & Gaussian Noise   & 0.9882 $\pm$ 0.0094 & 0.9803 $\pm$ 0.0186 & 0.0359 $\pm$ 0.0229   \\ 
 & Uniform Noise    & 0.9864 $\pm$ 0.0077 & 0.9774 $\pm$ 0.0138 & 0.0445 $\pm$ 0.0231 \\ 
 & SVHN             & 0.9522 $\pm$ 0.0120 & 0.9740 $\pm$ 0.0070 & 0.1610 $\pm$ 0.0349 \\ 
 & Textures         & 0.9087 $\pm$ 0.0092 & 0.8179 $\pm$ 0.0146 & 0.3054 $\pm$ 0.0291 \\ 
 & LSUN             & 0.9312 $\pm$ 0.0058 & 0.7818 $\pm$ 0.0166 & 0.2522 $\pm$ 0.0129 \\ 
 & TinyImageNet     & 0.8923 $\pm$ 0.0063 & 0.8747 $\pm$ 0.0073 & 0.3908 $\pm$ 0.0223  \\ 
 & Places365        & 0.9156 $\pm$ 0.0064 & 0.9702 $\pm$ 0.0024 & 0.3220 $\pm$ 0.0199 \\ \cline{2-5}
 & Average          & 0.9392 $\pm$ 0.0081 & 0.9109 $\pm$ 0.0115 &  0.2160 $\pm$ 0.0236 \\ \hline

\multirow{8}{*}{CIFAR-100}     
& Gaussian Noise & 1.0000 $\pm$ 0.0000 & 1.0000 $\pm$ 0.0000 & 0.0000 $\pm$ 0.0000 \\ 
& Uniform Noise  & 1.0000 $\pm$ 0.0000 & 1.0000 $\pm$ 0.0000 & 0.0000 $\pm$ 0.0001 \\ 
& SVHN           & 0.9701 $\pm$ 0.0015 & 0.9875 $\pm$ 0.0008 & 0.1534 $\pm$ 0.0067  \\ 
& Textures       & 0.8201 $\pm$ 0.0054 & 0.6665 $\pm$ 0.0077 & 0.5110 $\pm$ 0.0129  \\ 
& LSUN           & 0.8255 $\pm$ 0.0059 & 0.5350 $\pm$ 0.0112 & 0.5405 $\pm$ 0.0150  \\ 
& TinyImageNet   & 0.7880 $\pm$ 0.0029 & 0.7474 $\pm$ 0.0043 & 0.6113 $\pm$ 0.0029  \\ 
& Places365      & 0.8481 $\pm$ 0.0035 & 0.9420 $\pm$ 0.0015 & 0.4975 $\pm$ 0.0107 \\ \cline{2-5}
& Average        & 0.8931 $\pm$ 0.0027 & 0.8397 $\pm$ 0.0036 & 0.3305 $\pm$ 0.0069  \\ \hline
          
\multirow{8}{*}{SVHN}         
& Gaussian Noise & 0.9327 $\pm$ 0.0103 & 0.8245 $\pm$ 0.0303 & 0.2617 $\pm$ 0.0268  \\ 
& Uniform Noise  & 0.9204 $\pm$ 0.0086 & 0.7943 $\pm$ 0.0263 & 0.3113 $\pm$ 0.0237	\\ 
& CIFAR-100      & 0.9137 $\pm$ 0.0050 & 0.8018 $\pm$ 0.0093 & 0.3767 $\pm$ 0.0337  \\ 
& Textures       & 0.8917 $\pm$ 0.0054 & 0.6602 $\pm$ 0.0193 & 0.4515 $\pm$ 0.0346  \\ 
& LSUN           & 0.9206 $\pm$ 0.0050 & 0.5987 $\pm$ 0.0257 & 0.3335 $\pm$ 0.0313  \\ 
& TinyImageNet   & 0.9230 $\pm$ 0.0043 & 0.8169 $\pm$ 0.0109 & 0.3269 $\pm$ 0.0234  \\ 
& Places365      & 0.9242 $\pm$ 0.0065 & 0.9358 $\pm$ 0.0063 & 0.3146 $\pm$ 0.0299  \\ \cline{2-5}
& Average        & 0.9181 $\pm$ 0.0064 & 0.7760 $\pm$ 0.0183 & 0.3395 $\pm$ 0.0291  \\ \hline

\multirow{8}{*}{TinyImageNet} 
& Gaussian Noise    & 0.9967 $\pm$ 0.0020 & 0.9966 $\pm$ 0.0021 & 0.0149 $\pm$ 0.0103 \\ 
& Uniform Noise     & 0.9936 $\pm$ 0.0025 & 0.9935 $\pm$ 0.0026 & 0.0317 $\pm$ 0.0139 \\ 
& SVHN              & 0.9347 $\pm$ 0.0098 & 0.9720 $\pm$ 0.0047 & 0.3118 $\pm$ 0.0335 \\ 
& CIFAR-100         & 0.5999 $\pm$ 0.0035 & 0.5675 $\pm$ 0.0031 & 0.8534 $\pm$ 0.0040  \\ 
& Textures          & 0.6580 $\pm$ 0.0163 & 0.4754 $\pm$ 0.0272 & 0.7853 $\pm$ 0.0124  \\ 
& LSUN              & 0.6717 $\pm$ 0.0063 & 0.3382 $\pm$ 0.0055 & 0.7626 $\pm$ 0.0130  \\ 
& Places365         & 0.7022 $\pm$ 0.0061 & 0.8764 $\pm$ 0.0023 & 0.7472 $\pm$ 0.0109  \\ \cline{2-5}
& Average           & 0.7938 $\pm$ 0.0066 & 0.7457 $\pm$ 0.0068 & 0.5010 $\pm$ 0.0140  \\ \hline

\end{tabular}
\end{adjustbox}
\vspace{2mm}
\caption{Expanded results for the proposed method (i.e. MSP + norm-scaling).}
\label{tab:exp_odin}
\end{table*}}

\end{document}